# Identification of temporal transition of functional states using recurrent neural networks from functional MRI

Hongming Li, Yong Fan

Center for Biomedical Image Computing and Analytics,
Department of Radiology, Perelman School of Medicine, University of Pennsylvania

**Abstract.** Dynamic functional connectivity analysis provides valuable information for understanding brain functional activity underlying different cognitive processes. Besides sliding window based approaches, a variety of methods have been developed to automatically split the entire functional MRI scan into segments by detecting change points of functional signals to facilitate better characterization of temporally dynamic functional connectivity patterns. However, these methods are based on certain assumptions for the functional signals, such as Gaussian distribution, which are not necessarily suitable for the fMRI data. In this study, we develop a deep learning based framework for adaptively detecting temporally dynamic functional state transitions in a data-driven way without any explicit modeling assumptions, by leveraging recent advances in recurrent neural networks (RNNs) for sequence modeling. Particularly, we solve this problem in an anomaly detection framework with an assumption that the functional profile of one single time point could be reliably predicted based on its preceding profiles within stable functional state, while large prediction errors would occur around change points of functional states. We evaluate the proposed method using both task and resting-state fMRI data obtained from the human connectome project and experimental results have demonstrated that the proposed change point detection method could effectively identify change points between different task events and split the resting-state fMRI into segments with distinct functional connectivity patterns.

## 1 Introduction

Brain network analysis based on intrinsic functional connectivity (FC) derived from resting-state functional magnetic resonance imaging (fMRI) data enables us to investigate both static FC, estimated based on the entire fMRI scan, and dynamic FC, varying over the course of a fMRI scan [1, 2].

Existing studies of dynamic FC typically explore temporal dynamics based on network nodes defined by regions of interests (ROIs) based on anatomical atlases or functional data based brain parcellations, either using sliding-window (SW) based methods [2, 3] or splitting the entire fMRI scan into segments with quasi-static FC patterns [4-6]. In the SW methods, dynamic FC measures are estimated based on data points within multiple time windows, each of them with a fixed width but different starting positions shifted in time by a fixed number of data points. Notably, the SW methods' performance is hinged on the window parameters. Furthermore, it may not be an optimal way to use windows with a fixed width over the entire fMRI scan since the FC states may change at unpredictable intervals [7, 8]. To overcome limitations of the SW methods, a variety of methods have been developed to automatically split the entire fMRI scan into distinct segments, including Dynamic Connectivity Regression (DCR) methods [4], Bayesian inference based methods [6], Vector Autoregressive (VAR) model based methods [9], and statistical test based methods [5]. Different from the SW methods, these methods adaptively detect fMRI signal transitions to split the entire fMRI scan into segments. However, these methods are based on certain assumptions for the fMRI data, such as Gaussian distribution and VAR model, which are not necessarily well suited for fMRI data.

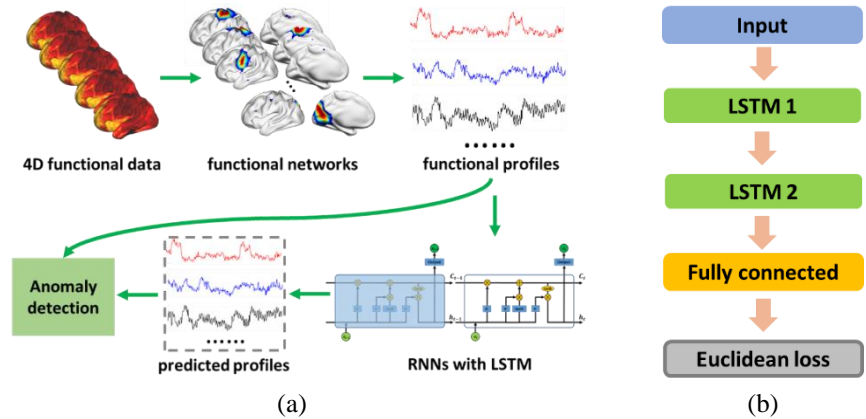

**Fig. 1.** Schematic illustration of our deep learning based change point detection framework. (a) The overall architecture of the proposed model, (b) the RNNs used in the model.

In this study, we develop a deep learning based framework for adaptively detecting dynamic functional state transitions in a data-driven way without any explicit model assumptions, by leveraging recent advances in deep learning based sequence modeling. Deep learning techniques, particularly recurrent neural networks (RNNs) with a long short term memory (LSTM) [10] structure, have achieved remarkable advances in sequence modeling [11], indicating that LSTM-RNNs might be suitable for characterizing fMRI data too. The basic assumption of the proposed deep learning based model is that the functional profile of one single time point could be reliably predicted based on its preceding profiles within a stable functional state, while large prediction errors would occur around change points of functional states. Given the predicted and real functional profiles, the change points are identified as anomaly time points with prediction errors larger than a predefined threshold value. We have applied the proposed method to both resting-state and task fMRI data obtained from the human connectome project (HCP) [12, 13], and experimental results have demonstrated that the proposed method could obtain better detection accuracy compared with state-of-the-art alternative methods on the task fMRI data, and also effectively detect change points that split the resting-state fMRI data into segments with significantly different functional connectivity patterns.

## 2   Methods

To identify temporal functional state transitions from fMRI data, recurrent neural networks (RNNs) with a LSTM structure [10] are trained based on functional profiles from a training cohort, where the functional profiles are extracted using a functional brain decomposition technique [14, 15]. Differences between the predicted functional profiles by the LSTM RNNs and the real ones on a validation cohort are then adopted to determine the optimal threshold for identifying the change points on the testing cohort. The overall framework is schematically illustrated in Fig. 1(a).

### 2.1   Prediction of functional profiles using LSTM RNNs

Given a group of $n$ subjects, each having a fMRI scan $X^i \in R^{T \times S}$, $i = 1,2,\dots,n$, consisting of $S$ voxels and $T$ time points, we first obtain $K$ functional networks $V^i \in R_+^{K \times S}$ and its corresponding functional time courses $U^i \in R^{T \times K}$ for each subject using a collaborative sparse brain decomposition method [14, 15] which could identify subject-specific functional networks with group level correspondence for better characterizing the intrinsic functional connectivity at an individual subject level. The functional time courses $U^i$, $i = 1,2,\dots,n$, are then used as training data to build a LSTM RNNs model for predicting functional profiles.

A LSTM RNNs model $M_{lstm}$ is built to predict the functional profile $U^i(t,\cdot)$ at each time point $t$ using its preceding functional profiles $\{U^i(t_p,\cdot), 1 \le t_p < t\}$ so that

$$U^i(t,\cdot) \approx \widetilde{U}^i(t,\cdot) = M_{lstm}(\{U^i(t_p,\cdot), 1 \le t_p < t\}). \tag{1}$$

Particularly, a LSTM RNNs model with 2 hidden layers is adopted, as shown in Fig. 1(b). Each hidden layer has 256 hidden nodes. A fully connected layer with $K$ output nodes is adopted for predicting the functional profiles. The Euclidean distance between real and predicted functional profiles is used as the objective function to optimize the RNNs model. We implement the model using Tensorflow [16].

### 2.2   Prediction based change point detection

Given the trained RNNs model $M_{lstm}$, we predict the functional profile for each time point $t$ ($t>1$) of every subject $i$, and the prediction error $E^i$ is measured by the deviation from its real functional profiles

$$E^i(t) = \left\| U^i(t,\cdot) - \widetilde{U}^i(t,\cdot) \right\|_2. \tag{2}$$

Assuming that the functional profiles could be reliably predicted for each time point based on its preceding functional signals within a quasi-stable functional state, we first detect the anomaly time points as those with relatively large prediction errors

$$A^i(t) = 1 \ if \ E^i(t) > T^i_v, and \ A^i(t) = 0 \ otherwise, \qquad (3)$$

where $A^i$ is the vector of length $T$ indicating that the $t$-th time point is one anomaly point if $A^i(t)$ equals to 1, and $T^i_v$ is the threshold value for identifying the predicted anomaly time points, to be determined as

$$T^i_v = mean(E^i) + \lambda * std(E^i), \qquad (4)$$

where $mean(x)$ and $std(x)$ denotes the mean and standard deviation of the vector $x$, $\lambda$ is a parameter used to adjust the threshold value.

Due to relatively low signal to noise ratio (SNR) of functional signals from fMRI data, the prediction errors evaluated at individual time points may oscillate a lot even for two consecutive time points. To improve the robustness and specificity of the identified change points, we apply a 1D convolutional operation to $E^i$ as

$$sE^i = conv1D(E^i, w(\sigma)), \qquad (5)$$

where $sE^i$ is a smoothed prediction error vector, $w$ is a Gaussian kernel with standard deviation $1/\sigma$, and a larger $\sigma$ corresponding to a narrower kernel. A change point is finally identified as the one with a local maximum $sE^i$ while its $E^i$ value is larger than the threshold $T^i_v$, i.e.,

$$C^i(t) = \begin{cases} 1, if \ A^i(t) = 1 \ and \ (sE^i, t) \ is \ a \ Local \ Maximum \\ 0, otherwise \end{cases}, \qquad (6)$$

where $C^i$ is the vector of length $T$ indicating that the $t$-th time point is one functional change point if $C^i(t)$ equals to 1.

## 3    Experimental results

We evaluated the proposed method based on both task and resting-state fMRI data of 490 subjects from the HCP [12, 13]. In this study, we focused on two tasks, including motor and working memory tasks. The motor task consisted of 6 events, including 5 movement events, namely left foot (LF), left hand (LH), right foot (RF), right hand (RH), tongue (T), and additionally 1 cue event (CUE) prior to each movement event. The working memory task consisted of 2-back and 0-back task blocks of tool, place, face and body, and a fixation period. The motor task fMRI scan of each subject contained 284 time points, while the working memory fMRI scan contained 405 time points. The resting-state fMRI scan of each subject contained 1200 time points. The fMRI data acquisition and task paradigm were detailed in [12, 13].

We applied the collaborative sparse brain decomposition method [14, 15] to the resting-state fMRI data of 490 subjects and identified 90 subject-specific functional networks (FNs) and their corresponding resting-state time courses. The number of FNs was automatically estimated by MELODIC of FSL [17]. The subject-specific FNs were then used to extract the time courses of task fMRI data for each subject. The proposed change point detection method was then applied to the motor task data, working memory data, and resting-state data respectively. Particularly, we split the whole dataset into training, validation, and testing datasets. The training dataset consisted of data of 400 subjects for training a LSTM-RNNs model for each task, the validation dataset included data of 50 subjects for selecting the optimal $\lambda$ and $\sigma$, and the testing dataset consisted of data of the remaining 40 subjects.

For the task fMRI data, the real change points were defined as the time points when each task event started or ended. The performance of change point detection was quantitatively evaluated using the distance between predicted change points and real ones. For each real change point, the distance to its nearest predicted change point was calculated, and the mean distance across all real change points was used to evaluate the sensitivity of the detection (error_sen). Moreover, the same measure was also calculated between each predicted change point and its nearest real change point to evaluate the specificity of the detection (error_spec). We have compared the proposed method with a Bayesian inference based method [6] in terms of their performance on the task fMRI data. As the Bayesian inference based method could achieve better performance on functional connectivity data with a relative small number of nodes, we picked up the motor and working memory related FNs (13 out of 90, and 24 out of 90 respectively) and applied the two change point detection methods to their functional profiles.

As no ground truth about change points is available on the resting-state fMRI data, two-sample covariance matrix testing [18] was adopted to examine if functional connectivity patterns of two consecutive data segments split by the detected change points were significantly different, and the differences were used as surrogate measures for evaluating the proposed method based on the resting-state fMRI data. The functional profiles of 90 FNs were used for change point detection on the resting-state fMRI data.

### 3.1 Change point detection on task fMRI data

We first selected the optimal parameters $\lambda$ and $\sigma$ using the validation dataset based on the error_sen and error_spec measures, as shown in Fig. 2(a) for the motor task fMRI data. Fig. 2(a, top) demonstrates that the error_sen decreased as $\sigma$ increased, a larger $\sigma$ corresponded a narrower smooth kernel, which led to noisy prediction error vectors and generated more change points. While generating more change points would improve the sensitivity of the detection, its specificity would decrease as shown in Fig. 2(a, bottom). The pattern of prediction errors in term of $\lambda$ had a similar trend as $\sigma$'s. We set $\sigma$ to 6, and $\lambda$ to 0 for the task fMRI data, taking into consideration both error_sen and error_spec, and applied the proposed method to the testing data.

The prediction performance on the motor task fMRI data of two randomly selected testing subjects are illustrated in Fig. 2(b). Most transitions between two consecutive task events were detected, and the identified change points were largely matched with the starting and ending time points of each task event. The overall prediction performance on the testing dataset is illustrated in Fig.2(c), our method obtained lower error_sen than the Bayesian method, and the error_spec was significantly lower (Wilcoxon signed rank test, $p < 0.05$).

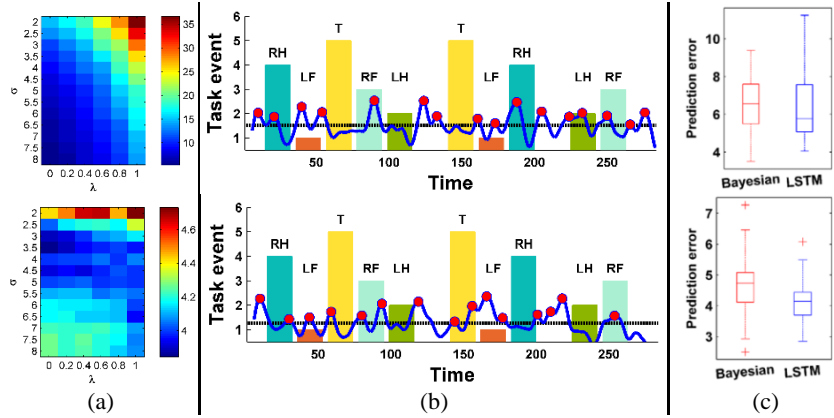

**Fig. 2.** Detection performance on motor task fMRI dataset. (a) Prediction errors on the validation dataset using different parameter settings, top: error_sen, bottom: error_spec, lower is better. (b) Identified change points of two randomly selected testing subjects: the $x$-axis denotes the time points, bar plots with different colors denote different task events ongoing at the located temporal interval, the blue curve is the smoothed prediction error, the dashed black line denotes the threshold used to identify the change points, and red circles denote the identified change points. (c) Prediction errors of 40 testing subjects, top: error_sen, bottom: error_spec.

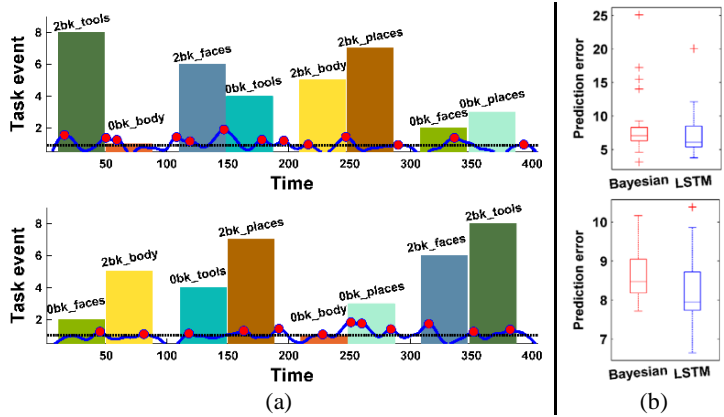

**Fig. 3.** Detection performance on the testing dataset of working memory task fMRI. (a) Identified change points of two randomly selected testing subjects, (b) Prediction error of 40 testing subjects, top: error_sen, bottom: error_spec.

The prediction performance on the testing dataset of working memory fMRI is illustrated in Fig. 3. The proposed method also obtained better performance on the working memory dataset than the Bayesian inference based method in terms of both detection sensitivity and specificity (Wilcoxon signed rank test, $p < 0.05$). We also evaluated our method based on the real change points adjusted by a hemodynamic lag of 6s for the task fMRI data, and our method outperformed the Bayesian inference based method.

### 3.2 Change point detection on resting-state fMRI data

We finally evaluated the proposed method using the testing dataset of resting-state fMRI data. As no ground truth about change points is available for selecting the optimal parameters $\lambda$ and $\sigma$, we set $\lambda$ to 1 and $\sigma$ to 3, aiming to detect a small number of change points and improve the prediction specificity. The identified change points on the resting-state fMRI data of one randomly selected testing subject are illustrated in Fig. 4 (top). The functional connectivity matrices of temporally dynamic segments between consecutive change points, as shown in Fig. 4(bottom), demonstrated that the functional connectivity patterns of consecutive segments were statistically significant (two-sample covariance matrix testing, $p < 0.05$), indicating that the change points detected by our method were functionally meaningful.

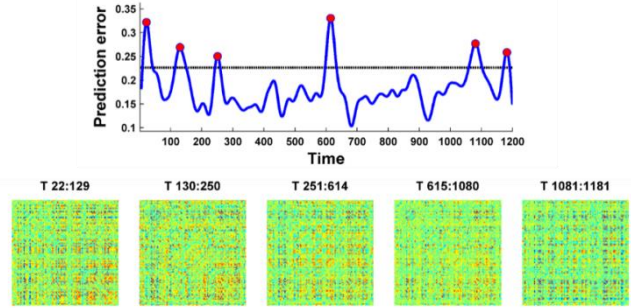

**Fig. 4.** Change point detection on resting-state fMRI data of one randomly selected testing subject. (top) the blue curve is the smoothed prediction error, the dashed black line denotes the threshold used to identify the change points, and red circles denote the identified change points, (bottom) the functional connectivity matrices of temporal segments split by change points.

### 4 Discussion and conclusions

We propose a LSTM RNNs based change point detection framework for identifying change points of temporal functional state transitions underlying different brain cognitive processes. Different from most of the existing change point detection methods, our learning based prediction model does not rely on any model assumption regarding the underlying functional profiles. The experimental results on the task fMRI data have demonstrated that our method could identify functionally meaningful change points with higher accuracy than a state-of-the-art method. The experimental results on the resting-state fMRI data further demonstrated that our method could effectively capture temporally dynamic functional states with distinct connectivity patterns.

### 5 Acknowledgements

This work was supported in part by National Institutes of Health grants [CA223358, EB022573, DK114786, DA039215, and DA039002] and a NVIDIA Academic GPU grant.

# References

1. Bullmore, E. and O. Sporns, *Complex brain networks: graph theoretical analysis of structural and functional systems.* Nat Rev Neurosci, 2009. **10**(3): p. 186-98.
2. Calhoun, V.D., et al., *The chronnectome: time-varying connectivity networks as the next frontier in fMRI data discovery.* Neuron, 2014. **84**(2): p. 262-274.
3. Hutchison, R.M., et al., *Dynamic functional connectivity: promise, issues, and interpretations.* Neuroimage, 2013. **80**: p. 360-78.
4. Cribben, I., et al., *Dynamic connectivity regression: determining state-related changes in brain connectivity.* Neuroimage, 2012. **61**(4): p. 907-20.
5. Jeong, S.O., C. Pae, and H.J. Park, *Connectivity-based change point detection for large-size functional networks.* Neuroimage, 2016. **143**: p. 353-363.
6. Zhang, J., et al., *Inferring functional interaction and transition patterns via dynamic Bayesian variable partition models.* Hum Brain Mapp, 2014. **35**(7): p. 3314-31.
7. Shakil, S., C.H. Lee, and S.D. Keilholz, *Evaluation of sliding window correlation performance for characterizing dynamic functional connectivity and brain states.* Neuroimage, 2016. **133**: p. 111-128.
8. Hindriks, R., et al., *Can sliding-window correlations reveal dynamic functional connectivity in resting-state fMRI?* Neuroimage, 2016. **127**: p. 242-256.
9. Samdin, S.B., et al., *A Unified Estimation Framework for State-Related Changes in Effective Brain Connectivity.* IEEE Trans Biomed Eng, 2017. **64**(4): p. 844-858.
10. Hochreiter, S. and J. Schmidhuber, *Long short-term memory.* Neural computation, 1997. **9**(8): p. 1735-1780.
11. Lipton, Z.C., J. Berkowitz, and C. Elkan, *A critical review of recurrent neural networks for sequence learning.* arXiv preprint arXiv:1506.00019, 2015.
12. Barch, D.M., et al., *Function in the human connectome: task-fMRI and individual differences in behavior.* Neuroimage, 2013. **80**: p. 169-89.
13. Glasser, M.F., et al., *The minimal preprocessing pipelines for the Human Connectome Project.* Neuroimage, 2013. **80**: p. 105-24.
14. Li, H., T.D. Satterthwaite, and Y. Fan, *Large-scale sparse functional networks from resting state fMRI.* Neuroimage, 2017. **156**: p. 1-13.
15. Li, H., T. Satterthwaite, and Y. Fan. *Identification of subject-specific brain functional networks using a collaborative sparse nonnegative matrix decomposition method.* in *2016 IEEE 13th International Symposium on Biomedical Imaging (ISBI)*. 2016.
16. Abadi, M., et al. *TensorFlow: A System for Large-Scale Machine Learning.*
17. Jenkinson, M., et al., *Fsl.* Neuroimage, 2012. **62**(2): p. 782-90.
18. Cai, T., W. Liu, and Y. Xia, *Two-sample covariance matrix testing and support recovery in high-dimensional and sparse settings.* Journal of the American Statistical Association, 2013. **108**(501): p. 265-277.